\documentclass[letterpaper,10pt,conference]{ieeeconf}
\IEEEoverridecommandlockouts
\usepackage{graphicx}
\usepackage{amsmath,amssymb}
\usepackage{booktabs}
\usepackage{array}
\usepackage{multirow}
\usepackage{cite}
\usepackage{xcolor}
\usepackage{url}
\usepackage{balance}
\usepackage{subcaption} 
\newcommand{\skillname}{Task2Hand}

\title{\LARGE \bf
Task2Hand: A Contact-Grounded Skill for\\
Parametric Dexterous-Hand Morphology Design
}

\author{Haoyang Wang$^{1}$, Weibo Gao$^{2}$, Zhengxiong Li$^{3}$,
\textit{Member, IEEE},\\ and Lingfeng Tao$^{4,*}$, \textit{Member, IEEE}%
\thanks{$^{1}$H. Wang is with the Department of Mechanical and Electrical Engineering,
Beijing University of Chemical Technology, Beijing 100029,
China (e-mail: vwang6925@gmail.com).}%
\thanks{$^{2}$W. Gao is with the Department of Mechanical and Aerospace Engineering,
North Carolina State University, Raleigh, NC 27695, USA (e-mail: wgao22@ncsu.edu).}%
\thanks{$^{3}$Z. Li is with the Department of Computer Science and Engineering,
University of Colorado Denver, Denver, CO 80217, USA (e-mail:
zhengxiong.li@ucdenver.edu).}%
\thanks{$^{4}$L. Tao is with the Department of Robotics and Mechatronics
Engineering, Kennesaw State University, Marietta, GA 30060, USA (e-mail:
ltao2@kennesaw.edu). $^{*}$Corresponding author.}%
}

\begin{document}
\maketitle
\thispagestyle{empty}
\pagestyle{empty}

\noindent\begin{abstract}
Dexterous hand design demands repeated expert work across geometry, robot-description code, collision debugging, grasp initialization, and policy training. We present \skillname{}, a \emph{Skill}: a specification package that turns a functional manipulation brief into a parameterized hand model and routes every candidate through cost-ordered contact, dynamics, and learning gates. Each gate returns a status and an evidence record; on failure, the Skill records the earliest blocker, revises one bounded parameter group, and reruns the cheaper gates. The agent generates, validates, and trains autonomously within declared bounds, and escalates trade-offs, plateaus, and conflicting evidence to a human who sets the next design direction. An evidence-driven design trajectory for bidirectional in-hand manipulation yielded SAGE Hand (\textbf{S}kill-\textbf{A}ware \textbf{G}enerative \textbf{E}mbodiment): four fingers, 16 actuated DoF, an integrated guiding palm, compact roots, cylindrical tips, and a collision-cleared opposable thumb. Under a matched training protocol, SAGE Hand reaches 45.6\% task success on a hexagonal prism against 37.3\% for the LEAP hand, and 70.8\% on a bicolor cylinder against 51.6\%. Dense simulated tactile input more than halves the cylinder drop rate. In a controlled iteration study, the Skill path preserves capture at 164/165 trials while the path without the Skill degrades to 14/165, confirming that the gates act as a regression guard during refinement. Accepted candidates pack with URDF/MJCF, metrics, checkpoints, and a provenance manifest.
\end{abstract}

\section{Introduction}

\noindent A robotic dexterous hand's morphology sets the limits of what any controller can do with it. Finger length, root placement, thumb opposition, joint axes, fingertip geometry and palm shape together decide whether contacts can be made, adjusted and remade before the object falls~\cite{bicchi2000hands}. No learning algorithm recovers a contact transition the mechanism forbids. Building such a hand requires time-consuming expert work. The designer cycles between a geometric concept, a URDF or MJCF implementation, collision debugging, a stable initial grasp, dynamic tests, and reinforcement learning (RL). A model that looks reasonable can still hide a misoriented axis, self-interference, a narrow grasp basin, or an initial state that holds only because two bodies overlap. These faults particularly affect users who know their manipulation requirements but are not specialists in hand design.

Language is a natural interface here, though words alone do not establish mechanical validity. What matters is the supporting workflow: parameterized source code, physical tests, a fixed learning protocol, and a result package. We design this workflow as an agent tool: \emph{Task2Hand Skill}~--- an instruction file plus generators, validators, and evidence schemas that a general-purpose coding agent follows.

\begin{figure}[t]
  \centering
  \includegraphics[width=\columnwidth]{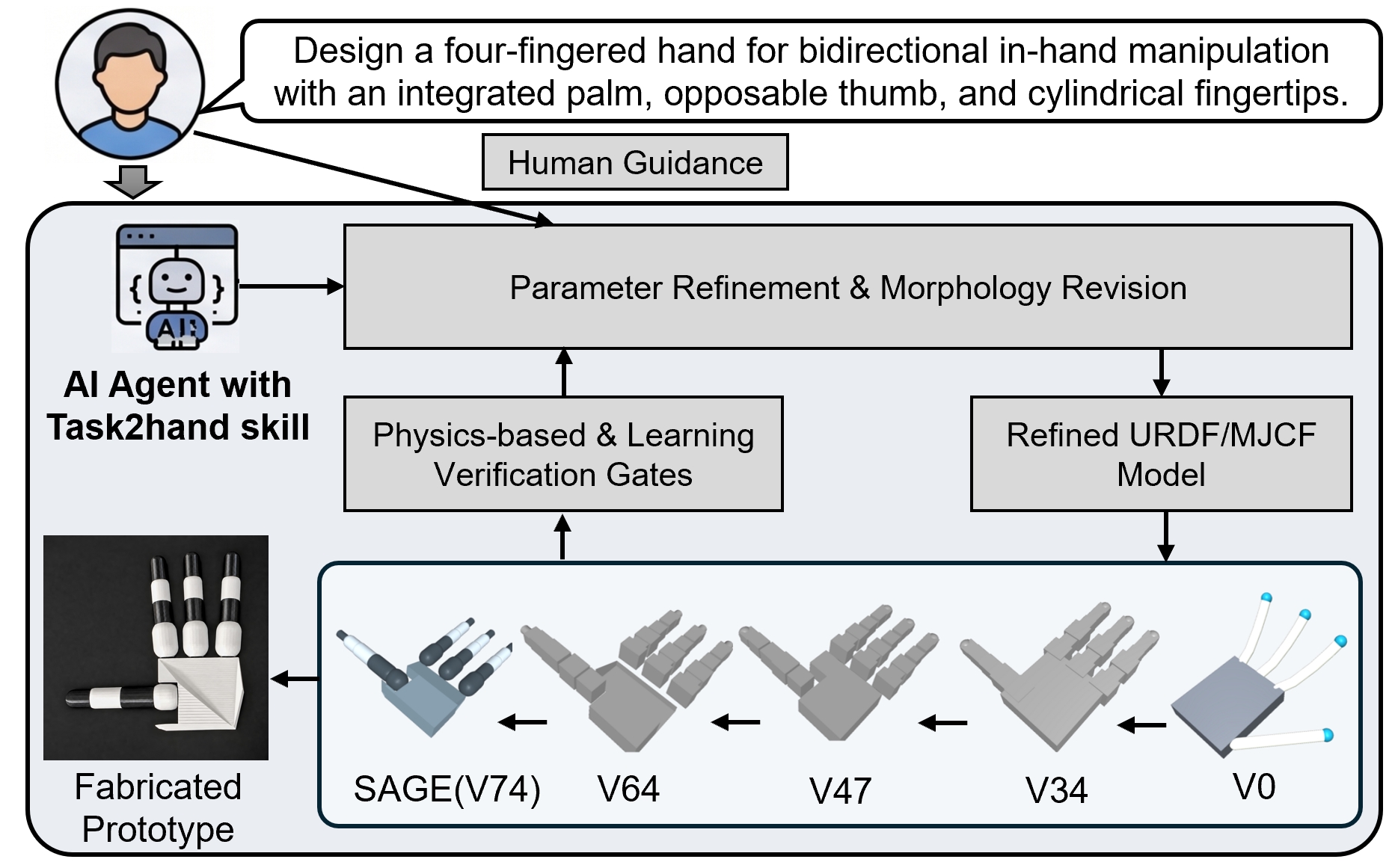}
  \caption{Overview of the \skillname{} design loop. A human provides a functional manipulation brief and design guidance, and an AI coding agent equipped with the \skillname{} Skill generates parameterized hand-model candidates. Physics-based and learning verification gates evaluate each candidate, and failure evidence guides bounded parameter refinement and morphology revision. The design history depicts the resulting evolution toward the final design and a fabricated prototype.}
  \label{fig:1}
\end{figure}

The gates split the loop by each side's responsibilities. The agent handles what is checkable by machine, including generation, validation, bounded search, training, and analysis, and runs autonomously within declared bounds. When the evidence exposes a trade-off, a plateau, or conflicting signals across gates, the Skill escalates to a human who sets the next design direction. Every escalation and answer is written to the ledger, so the design path is replayable. We characterize the boundary between autonomous and human-guided operation empirically in Sec.~\ref{section:method}.

Three questions follow. \textbf{RQ1}: Can an AI coding agent turn a non-expert brief into an executable, physically admissible hand by way of reusable gates, and do those gates protect validated capabilities during subsequent iteration? \textbf{RQ2}: Does the morphology accepted through the interactive loop achieve higher rotation success under a fixed, matched protocol? \textbf{RQ3}: Which conclusions hold across objects, and does temporal tactile input change them?

Our test case is bidirectional in-hand manipulation, which demands signed rotation authority, object retention, coordinated contact release, and lateral fingertip motion. Starting from a four-finger layout, a user named functional problems and a coding agent wrote every model, validator, search script, training interface, and analysis artifact. The resulting design process produced a task-specialized dexterous hand, which we call SAGE Hand (\textbf{S}kill-\textbf{A}ware \textbf{G}enerative \textbf{E}mbodiment), together with the failure cases from which we distilled the Skill (Fig.~\ref{fig:1}).

Our contributions are:
\begin{itemize}
  \item a reusable \skillname{} contract that maps a natural-language requirement and a bounded parameter space to URDF/MJCF, contact-grounded gates, RL packaging, and a provenance ledger;
  \item an interaction protocol with named escalation conditions, at most three costed options per escalation, and a recorded answer, together with an empirical characterization of the boundary between autonomous and human-guided operation;
  \item an evidence-driven design trajectory for bidirectional in-hand rotation that yielded SAGE Hand: a four-finger dexterous hand with 16 actuated DoF, an integrated guiding palm, compact roots, cylindrical tips, and a collision-cleared opposable thumb;
  \item a matched evaluation on three objects in which SAGE Hand achieves a higher success rate than the reference hands, simulated tactile input more than halves the cylinder drop rate, and a controlled iteration study showing that the Skill path preserves capture at 164/165 trials while the path without the Skill degrades to 14/165.
\end{itemize}

Given a specialized Skill and a bounded number of human decisions at evidence-defined points, an agent can carry a contact-rich morphology design from a brief to a hand model with a trained controller and auditable design records.

\section{Related Work}

\subsection{Dexterous Hands and In-Hand Manipulation}
\noindent Rigid direct-drive hands such as Allegro~\cite{allegro2026spec} and LEAP~\cite{shaw2023leap} provide compact research platforms, whereas Shadow, ORCA, and DLR/HIT employ more anthropomorphic, tendon-driven, or modular designs~\cite{shadow2024spec,christoph2025orca,dlrhit22026}. RBO Hand~3 instead uses intrinsic compliance to adapt contact geometry~\cite{puhlmann2022rbo}. These platforms span a wide range of actuation, palm, finger, and sensing choices, yet each fixes its morphology before any policy is optimized on it. Learning-based systems have produced coordinated contact behavior and finger gaiting~\cite{andrychowicz2020dexterous}, tactile observations can drive rotation without vision~\cite{yin2023rotating}, and multimodal skins read normal, shear, and thermal signals~\cite{guo2025bioskin}. We work one step earlier, on revising the morphology itself while the policy is being evaluated.

\subsection{Task-Conditioned Robot Design}
\noindent Differentiable design~\cite{xu2021contact}, graph-grammar search~\cite{zhao2020robogrammar}, and language-conditioned systems~\cite{ringel2025text2robot,khan2024text2cad} connect tasks to robot morphology through different representations. Generative and evolutionary methods have also been applied to end-effectors and embodiments: Fit2Form learns a generative model of gripper geometry conditioned on objects~\cite{ha2021fit2form}, and DERL co-evolves morphology and control at scale, showing that evolved bodies subsequently learn faster~\cite{gupta2021derl}. Closer to our setting, RoboMorph places an LLM mutation operator inside an evolution-and-RL loop~\cite{qiu2026robomorph}; Lang2Morph maps language to manufacturable hand geometry~\cite{qiao2025lang2morph}; and House of Dextra studies morphology--policy co-design across embodiments~\cite{fay2026houseofdextra}.

Each of these supplies a search operator over a design space. \skillname{} is complementary: it fixes the \emph{acceptance criteria} that decide which proposal survives. Those criteria are grounded in simulator contact and learning outcomes, so a candidate is judged by whether it can be grasped, controlled, and trained, not by whether it looks plausible. The failure ledger belongs to the method and the contract lives in the filesystem, so any coding agent that implements a search operator can plug into the same gates.

\begin{figure*}[t]
  \centering
  \includegraphics[width=\textwidth]{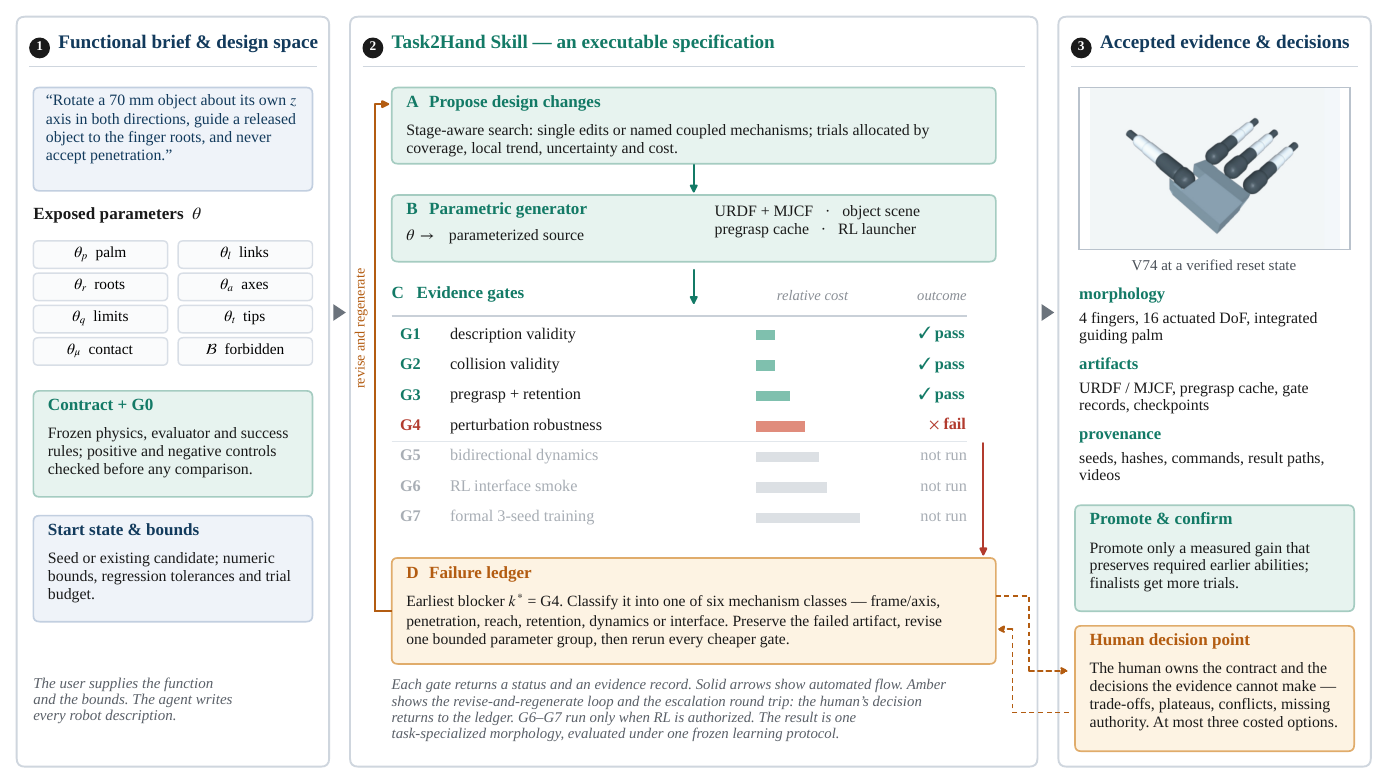}
  \caption{\textbf{The \skillname{} Skill.} \textbf{(1)} The user writes a
  functional brief and exposes the parameter set $\theta$ the agent may change;
  the contract, evaluator, and success rules are frozen before any comparison.
  \textbf{(2)} The Skill proposes one bounded edit, regenerates the whole model
  from $\theta$, and runs the cost-ordered gates \mbox{G1--G7}. Every gate
  returns a status and an evidence record, so the earliest blocker $k^{*}$ names
  a failing mechanism and the ledger identifies a parameter group for targeted revision.
  One group is revised, and every cheaper gate is rerun. Where the evidence poses
  a choice it cannot settle, the Skill escalates with at most three costed
  options; the decision returns to the ledger and the search resumes inside the
  approved scope. \textbf{(3)} An accepted candidate carries its artifacts and
  provenance. In the iteration shown $k^{*}=\mathrm{G4}$, so \mbox{G5--G7} never
  ran and no training budget was spent.}
  \label{fig:skill}
\end{figure*}

\section{The \skillname{} Skill}
\label{section:method}

\subsection{Human Contract and Parameterized Representation}
\noindent A Skill (Fig.~\ref{fig:skill}) is a package on disk, not a prompt. Its instruction file fixes the workflow. Generators produce isolated candidates, validators enforce the task and candidate schemas, and evidence records tie every decision back to models, metrics, and videos. We write a task contract as
\begin{equation}
\mathcal{C}=\{\mathcal{T},\mathcal{O},\mathcal{F},\mathcal{B},\mathcal{P}\},
\end{equation}
where $\mathcal{T}$ is the target manipulation, $\mathcal{O}$ specifies the object and release distribution, $\mathcal{F}$ lists parameters the agent may change, $\mathcal{B}$ lists forbidden states and workarounds, and $\mathcal{P}$ is the frozen evaluation protocol. The contract is written once, before any candidate exists, and it is what later decisions are checked against. The exposed morphology vector is
\begin{equation}
\boldsymbol{\theta}=\{\theta_{p},\theta_{l},\theta_{r},\theta_{a},\theta_{q},\theta_{t},\theta_{\mu}\},
\end{equation}
whose terms are palm profile, link dimensions, root poses, joint axes, joint limits, tip geometry, and contact/friction parameters. A user changes functional quantities and never touches an individual robot-description element. What comes back is more than geometry: parameterized source, URDF and MJCF, the object scene, a pregrasp cache, gate records, an RL launcher, checkpoints, metrics, videos, and a manifest of seeds, hashes, commands, and result paths.

\subsection{Human Role and AI Role}
\label{sec:humanrole}
\noindent The Skill splits the loop by what each side can be held to. The agent owns everything checkable by machine: it writes generators, robot descriptions, validators, search scripts, training interfaces, and analysis artifacts, proposes bounded edits, and runs the gate cascade. The human owns what fixes meaning: the contract $\mathcal{C}$, the bounds $\mathcal{B}$, the acceptance rules, and the design decisions the evidence cannot settle. The human states the function and never edits a robot element by hand.

Between escalations, the agent runs unsupervised inside an approved scope~--- a parameter group, numeric bounds, a regression tolerance, and a trial budget. It escalates in four situations: a \emph{trade-off}, where clearing the blocking gate would regress an earlier one; a \emph{plateau}, where no proposal in the budget is promotable; \emph{conflicting evidence} across gates or repeats; and a move that falls outside the authority it was given. An escalation carries the current candidate, the earliest blocker with its evidence, and at most three options with expected cost and regression risk. The human answers with one decision~--- choose an option, widen or narrow a bound, authorize learning, or stop~--- and the answer is written to the ledger with the state it applied to. The human reviews renderings, videos, and metrics; the physical gates carry the verification burden that code review would otherwise carry.

A general-purpose AI coding agent, running in the Codex desktop application between July and September 2026, wrote every generator, validator, search script, training interface, and analysis artifact in this project. The auditable R7 run requested model \texttt{gpt-5.6-sol} at high reasoning effort; the runtime identified itself only as Codex with a GPT-5-based model. The study evaluates the Skill and the human--agent division of labour, holding the coding agent fixed.

\subsection{Progressive Evidence Gates}
\noindent A gate maps a candidate to a status and an evidence record,
\begin{equation}
G_k:\;(\boldsymbol{\theta};\mathcal{C})\longmapsto(s_k,E_k),\qquad s_k\in\mathcal{S},
\end{equation}
with $\mathcal{S}=\{\mathrm{pass},\mathrm{fail},\mathrm{diagnostic\text{-}only},\mathrm{blocked},\mathrm{not\text{-}run}\}$ and $E_k$ the trials, seeds, hashes, metrics, and videos behind $s_k$. A diagnostic run~--- collision disabled, object teleported, threshold relaxed~--- can locate a cause, but it does not advance a candidate. The seven gates are:
\begin{enumerate}
  \item \textbf{Description validity:} parse topology, frames, limits, inertias, actuators, and finite simulator state.
  \item \textbf{Collision validity:} reject self-interference and any accepted object-hand state with negative contact distance.
  \item \textbf{Pregrasp and static retention:} release the object above the hand, close slowly, and require a stable nonpenetrating RL initial state.
  \item \textbf{Perturbation robustness:} perturb joints and object within bounds to expose narrow capture basins or hidden overlaps.
  \item \textbf{Bidirectional dynamics:} test both yaw signs, retention, drift, contact switching, and numerical stability.
  \item \textbf{RL interface smoke:} verify observations, actions, reward, MPI updates, checkpoint serialization, and teacher-free evaluation.
  \item \textbf{Formal evidence:} run the frozen multi-seed protocol and retain best and final checkpoints separately.
\end{enumerate}
MuJoCo settles all contact dynamics~\cite{todorov2012mujoco}; URDF renderings serve for geometry and kinematic review only. A candidate is ready for formal training when
\begin{equation}
R(\boldsymbol{\theta};\mathcal{C})=\bigwedge_{k=1}^{6}\left[s_k=\mathrm{pass}\right],
\end{equation}
and acceptance also requires $s_7=\mathrm{pass}$. We call
\begin{equation}
k^{*}(\boldsymbol{\theta})=\min\{k:\,s_k\neq\mathrm{pass}\}
\label{eq:blocker}
\end{equation}
the earliest blocker and leave every later gate unclaimed, so a candidate reports capability only up to $k^{*}-1$.

The gate order follows from cost, subject to dependency constraints. Let $c_k$ be the expected cost of running gate $k$ and $q_k=\Pr[s_k=\mathrm{pass}\mid s_{1:k-1}=\mathrm{pass}]$ its conditional pass rate. Screening one candidate under an ordering $\pi$ costs
\begin{equation}
\mathbb{E}\!\left[C(\pi)\right]=\sum_{k=1}^{7}c_{\pi(k)}\prod_{j<k}q_{\pi(j)},
\label{eq:cost}
\end{equation}
which is minimized over unconstrained orderings by sorting on nondecreasing $c_k/(1-q_k)$: cheap, frequently rejecting tests first. A gate can only run on the state its predecessors produce, which imposes a partial order $\prec$: $G_1\prec G_2$ because contact distances are undefined on a description that does not parse, $G_3\prec G_4$ and $G_3\prec G_5$ because both perturbation and dynamics act on a validated initial state, and $G_6\prec G_7$ because training needs a verified learning interface. We minimize~\eqref{eq:cost} over the linear extensions of $\prec$; the resulting ordering is frozen before the study.

$G_5$ illustrates the interplay between cost and dependency. It carries the lowest conditional pass rate of the seven~--- 203 of the 209 physics-trial failures in Sec.~\ref{sec:auto} fall on it~--- so on $c_k/(1-q_k)$ alone it would move early, but every bidirectional trial consumes a validated initial state and a full episode of contact dynamics, and it cannot precede $G_3$. The cascade therefore spends its cheap gates first and reaches the expensive, discriminative one only on candidates that are already admissible. Placing learning last follows the same logic: formal training evaluates only candidates whose contact, reset, and control interface are already verified.

The cascade keeps initialization and control apart. Writing $p_{\mathrm{init}}(\boldsymbol{\theta})$ for the probability of forming a legal, retained, nonpenetrating initial state and $p_{\mathrm{ctrl}}(\boldsymbol{\theta})$ for the probability of task success given such a state, the end-to-end rate is
\begin{equation}
p_{\mathrm{task}}(\boldsymbol{\theta})=p_{\mathrm{init}}(\boldsymbol{\theta})\,p_{\mathrm{ctrl}}(\boldsymbol{\theta}).
\label{eq:factor}
\end{equation}
Gates $G_3$ and $G_4$ bound $p_{\mathrm{init}}$ before any policy is trained, and $G_7$ estimates $p_{\mathrm{ctrl}}$ under the protocol. The two factors call for different revisions, so we report them separately. Each gate also returns a scalar margin $m_k(\boldsymbol{\theta})$ in its own units: the minimum signed contact distance for $G_2$, the fraction of legal retained captures for $G_3$, the surviving fraction under bounded perturbation for $G_4$, the achieved yaw amplitude in the weaker signed direction for $G_5$, and late-training success for $G_7$. The per-gate tolerance $\varepsilon_{k^{*}}$ in Sec.~\ref{sec:revision} sets the minimum improvement required for promotion.

\subsection{Failure Classification and Bounded Revision}
\label{sec:revision}
\noindent Every blocked candidate is filed by cause. A classifier
\begin{equation}
\kappa:\;\left(k^{*},E_{k^{*}}\right)\longmapsto f\in\mathcal{F},
\end{equation}
maps the earliest blocker and its evidence onto $\mathcal{F}=\{$frame/axis, penetration, reach, retention, dynamics, interface$\}$, and a fixed table sends each cause to the parameter group that can answer it. The revision is deliberately narrow,
\begin{equation}
\boldsymbol{\theta}'=\Pi_{\mathcal{B}}\!\left(\boldsymbol{\theta}+\boldsymbol{\delta}\right),\qquad\mathrm{supp}(\boldsymbol{\delta})\subseteq g(f),
\label{eq:revise}
\end{equation}
where $g(f)$ is that single group~--- or one named coupled mechanism, declared in advance~--- and $\Pi_{\mathcal{B}}$ projects onto the declared bounds. The failed artifact is preserved, so a revision adds to the history instead of editing it.

A revised candidate must re-earn what its predecessor had. It replaces the incumbent only if
\begin{equation}
k^{*}(\boldsymbol{\theta}')>k^{*}(\boldsymbol{\theta})\ \text{or}\ \left[k^{*}(\boldsymbol{\theta}')=k^{*}(\boldsymbol{\theta})\wedge m(\boldsymbol{\theta}')>m(\boldsymbol{\theta})+\varepsilon\right]
\label{eq:promote}
\end{equation}
with $m_{k^{*}}$ the margin at the blocking gate and $\varepsilon_{k^{*}}$ the tolerance declared for it, and only after every gate below $k^{*}$ has been rerun and passed again.

Two properties follow. First, a sequence built this way is not a single-factor experiment: consecutive candidates differ by the revision the evidence asked for, so each local decision can be read, while the endpoint cannot be attributed to any one change. Second, when~\eqref{eq:promote} fails for a budgeted number of consecutive proposals, no improving bounded move is found within the trial budget. That is the plateau condition of Sec.~\ref{sec:humanrole}, and it is where the human enters. Table~\ref{tab:traces} gives three traces from the V-series.

\section{SAGE Hand Case Study and Experimental Protocol}

\noindent This section instantiates the Skill on one task. The brief required signed Rotate-Z motion, passive guidance of a released object toward the finger roots, lateral contact renewal, and rejection of every accepted object--hand state with geometric penetration. Seventy-four candidates were generated under that contract; Table~\ref{tab:traces} reads three of the transitions in the format of Sec.~\ref{sec:revision}, each showing the earliest blocker, the bounded revision it licensed, and the evidence preserved for it. The rest of this section describes the accepted hand, the objects, the hands it is compared against, and the frozen protocol under which every number in Sec.~\ref{sec:results} was produced.

\begin{table*}[t]
\captionsetup{font=small}
\caption{Representative evidence-driven morphology traces from the V-series.}
\label{tab:traces}
\centering
\footnotesize
\begin{tabular}{p{0.12\textwidth}p{0.22\textwidth}p{0.25\textwidth}p{0.31\textwidth}}
\toprule
Trace & Earliest blocker & Bounded revision & Preserved evidence and interpretation \\
\midrule
Prototype P2$\rightarrow$Prototype P3 & Mount/reset convention obscured signed motion & Correct the frame and reset before changing geometry & Scripted motion reached $+15.13^{\circ}$ and $-13.87^{\circ}$: both signed directions were physically available. \\
Prototype P4$\rightarrow$Prototype P5 & Object drifted outward; ramp candidates penetrated or lost contact & Replace the added shelf with one continuous guiding palm surface & Invalid ramps were rejected; the retained profile redirected free-fall toward the finger roots without a separate intersecting part. \\
V72$\rightarrow$SAGE Hand & Thumb-root interference blocked lateral contact transfer & Recess the palm root, reposition the thumb axis, keep physical collision & Captures rose from 0/165 to 14/165 without disabling the collision pair. \\
\bottomrule
\end{tabular}
\end{table*}

\subsection{Task-Derived Morphology}
\noindent SAGE Hand has four fingers and 16 actuated DoF. Its palm is one integrated body, and the forward guiding profile on that body pushes back against motion toward the wrist while steering a released object down to the finger roots. Link lengths are moderate, which keeps reach without crowding the middle of the workspace. Cylindrical tips present a longer lateral contact patch than spherical alternatives. Fanning the root axes widens the side-push workspace. The thumb is shorter, repositioned, and turned onto a palm-normal lateral axis, and a physical recess at its root lets it push or pull across the side of the object with its collision geometry intact.

Together, the hand can rotate without keeping all four fingertips down. Thumb push and pull, outer-finger restraint, and contact renewal can alternate. The design was a large set of valid contact transitions, not a static four-point cage.

\subsection{Three Representative Objects}
\begin{figure}[t]
  \centering
  \includegraphics[width=\columnwidth]{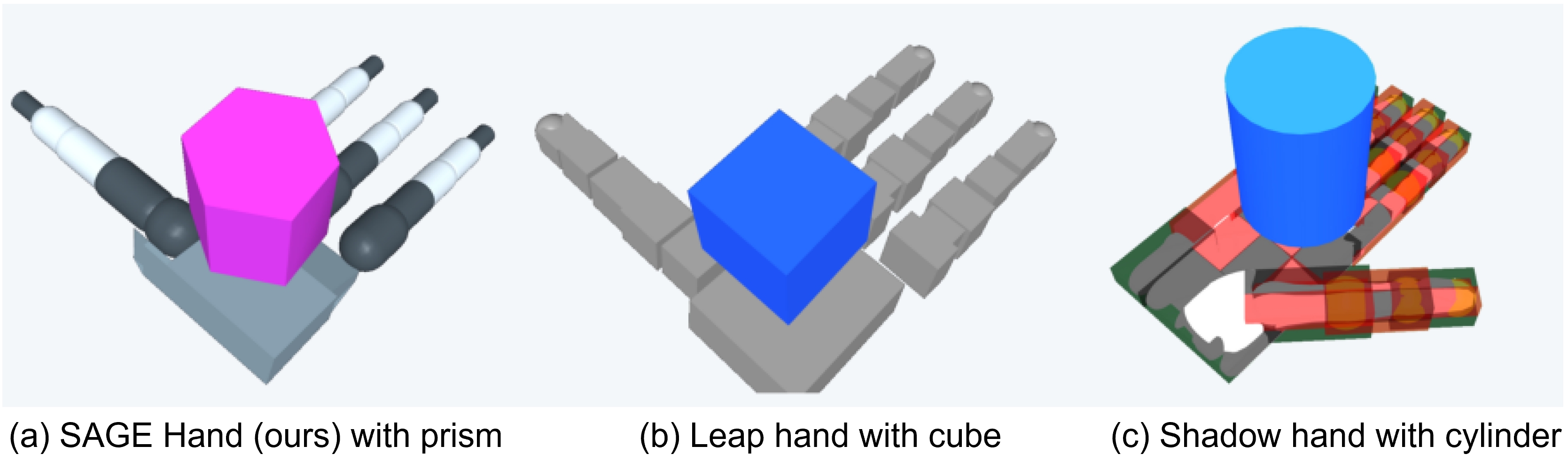}
  \caption{\textbf{Evaluation setup.} Three hands at a verified reset state: (a) SAGE Hand, the morphology the Skill produced, with the hexagonal prism; (b) the LEAP hand with the cube; (c) the Shadow hand with the cylinder. Every hand is run on all three objects under the same physics, task definition, reward, and training budget.}
  \label{fig:exp}
\end{figure}

\noindent The asymmetric hexagonal prism has a unique orientation and discontinuous faces. The colored cube gives the familiar six-face orientation task. The bicolor cylinder sits with its axis normal to the palm, which makes rotation easy to see, and its continuous side surface makes contact migration easy. Across the three (Fig.~\ref{fig:exp}), we can ask whether a mechanism seen on one contact geometry transfers to another.

\begin{table}[t]
\captionsetup{font=small}
\caption{Experimental hands. SAGE Hand versus LEAP is a morphology comparison on matched action spaces; SAGE Hand~T4 versus SAGE Hand is a sensing ablation on identical collision geometry; Shadow serves as an anthropomorphic contextual reference.}
\label{tab:hands}
\centering
\footnotesize
\setlength{\tabcolsep}{4pt}
\begin{tabular}{lccp{0.40\columnwidth}}
\toprule
Hand & Act.\ DoF & Tactile & Scientific role \\
\midrule
SAGE Hand & 16 & 0 & Agent-generated morphology; primary output \\
SAGE Hand T4 & 16 & 74 & SAGE Hand with tactile sensing \\
LEAP hand & 16 & 0 & Baseline from Public LEAP Hand model \\
Shadow hand & 18 & 92 & Anthropomorphic contextual reference \\
\bottomrule
\end{tabular}
\end{table}

\subsection{Hands and Controlled Comparisons}
\noindent Table~\ref{tab:hands} keeps the morphology comparison and the sensing ablation apart. SAGE Hand and the LEAP hand share a 16-action, no-tactile interface. SAGE Hand T4 keeps SAGE Hand's collision geometry and adds 74 simulated tactile regions with temporal contact features. The Shadow hand is wrist-locked, tilted by $10^{\circ}$, and uses 18 actions on a different morphology; it provides anthropomorphic context rather than a controlled contrast.

The SAGE Hand--LEAP contrast measures task specialization: SAGE Hand is the endpoint of 74 revisions aimed at this task, and the reference had none. The reference is a project reconstruction from the released LEAP model.

The pregrasp budgets differ. Each worker validates one morphology-specific start. SAGE Hand does a staged free drop and finger closure; the LEAP hand restores an offline cache and holds it for 500 substeps at reset. Both validated states contain three contacting hand geoms, and the search that produced them is on record. SAGE Hand screened 165 deterministic closure controls after a 3-mm free drop, found 14 that were functional and nonpenetrating, and took \texttt{local\_067}. The LEAP vertical-prism builder scanned up to $21\times21\times7=3{,}087$ object offsets offline and took the first to survive a $1{,}000$-substep hold.

\subsection{Reset, Reward, and Learning}
\noindent For SAGE Hand the object is released from a fixed morphology-specific pose with roughly a 3-mm surface gap, falls under gravity, and the fingers close slowly onto a validated cached state. Reference hands restore their own independently validated cache before policy control begins. Neither initialization enters replay or reward. Each environment process builds its cache once at startup, and if the build fails, initialization aborts. During an episode, every candidate transition is checked at each physics substep and rolled back before an illegal object--finger overlap can be accepted.

The MuJoCo time step is 0.002~s and each action spans 20 substeps, giving 25-Hz control and a 100-action (4-s) episode. Dynamic hand/object friction is $(1.6,0.015,0.001)$ and floor friction is $(1.2,0.02,0.001)$ in MuJoCo's three friction coordinates. Every object has mass 0.1~kg. The penetration tolerance is 0~m: any reported contact distance below zero triggers rollback.

The relative target rotation is sampled uniformly over $[-180^{\circ},180^{\circ}]$. Full quaternion error is
\begin{equation}
e_R=2\arccos\left(\left|\langle q_o,q_g\rangle\right|\right),
\end{equation}
and a trial succeeds when $e_R<5^{\circ}$ and the object is still held. Because $e_R$ measures full orientation error, accumulated tilt can fail a trial even if the yaw target was met. Reward is 0 for success, $-1$ for ordinary failure, and $-2$ for a drop or illegal contact. Shadow carries one extra condition (target-axis speed below $0.15$~rad/s for 20 control steps).

Target-conditioned policies use TD3~\cite{fujimoto2018td3} with hindsight experience replay~\cite{andrychowicz2017her}. Every sampled transition stays eligible for future-goal relabelling, including those that earned $-2$. Drop height and the illegal-contact flag are preserved in the achieved-goal state, so a dropped or illegal transition still scores $-2$ after relabelling. Formal runs use 400 epochs, 50 cycles per epoch, 40 update batches per cycle, 10 MPI ranks, and seeds 772, 882, and 992. Every five epochs we evaluate 100 teacher-free episodes. For the morphology-evolution analysis, we use the mean of the last ten available evaluations (epochs 355--400) to characterize late-training performance. For the final cross-hand comparison, we report the mean of the final ten available evaluations as the primary metric, while retaining the best checkpoint from each seed as an auxiliary best-observed result for deployment and qualitative evaluation. Jobs ran on a 24-core i9-12900KF server with 31~GiB RAM and an RTX 3090; mean wall time was 13.25~h per 400-epoch.

\subsection{Statistical Treatment}
\noindent Unless otherwise noted, each cell aggregates three seeds. Tables give mean $\pm$ between-seed SD; text gives between-hand differences with two-sided Welch $t$ intervals. An effect is \emph{established} when its 95\% interval excludes zero and \emph{directional} when it does not. 

We analyze the two colored-object packages separately. The primary evaluation package uses a replay capacity of $10^{6}$ episodes; the 09-01 package uses $4\times10^{5}$, float32 replay storage, and different hardware. The two packages share seeds 772 and 992, so agreement between them tests sensitivity to replay capacity and hardware.

\begin{figure}[t]
  \centering
  \includegraphics[width=\columnwidth]{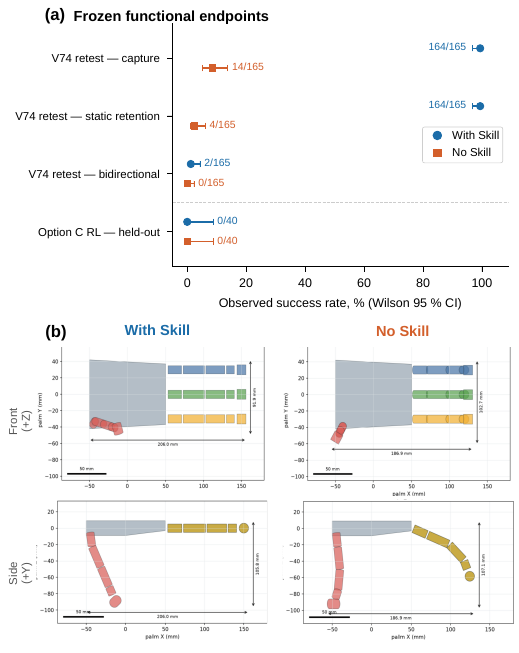}
  \caption{\textbf{Fully automatic iteration from V74 with and without the Skill.} \textbf{(a)}~Frozen functional endpoints (Wilson 95\,\% CI). The Skill path preserves capture (164/165) and static retention (164/165); the path without the Skill degrades both to 14/165 and 4/165. Neither path reaches held-out RL success under a 50-epoch budget. \textbf{(b)}~Front and side views of the two resulting morphologies at $q_{\mathrm{open}}$, unified camera and scale. The Skill-iterated hand retains flat finger geometry; the No-Skill hand introduces distal finger droop consistent with the capture-rate gap in~(a).}
  \label{fig:search}
\end{figure}

\section{Results}
\label{sec:results}

\subsection{Fully Automatic Operation and Regression Guard}
\label{sec:auto}

\noindent We evaluate the Skill on two axes: how far automatic operation proceeds from scratch, and whether the gates protect validated capabilities during iteration (Fig.~\ref{fig:search}).

\textit{From-scratch generation.} Both a Skill-guided path and six randomized plain-search paths produce parseable, collision-clean hands and clear the cheap gates. The Skill path is deterministic: one run, legal capture $95.7$\% and static retention $95.6$\%. Five of six plain-search seeds reach the same level; one collapses to $48.4$\% and $45.9$\%. Both arms stop at the same gate: no path exceeds $6$\% in $+Z$, and every path is essentially zero in $-Z$, so no path passes the bidirectional-dynamics gate. Of 209 failures in a fixed-budget audit, 203 fall at $G_5$, and no training is spent locating the bottleneck.

\textit{Regression guard under iteration.} Starting from V74, we run one bounded-iteration cycle with the Skill and one without, under matched protocols. The Skill path preserves capture at 164/165 trials and static retention at 164/165; without the Skill both degrade to 14/165 and 4/165. Neither path achieves held-out RL success under a 50-epoch budget (0/40), but their failure modes differ: the Skill hand triggers the contact guard in 21/40 episodes, indicating persistent contact-transition attempts, while the No-Skill hand avoids contact change and times out in 25/40. Fig.~\ref{fig:search}(b) shows the geometric cause: the Skill-iterated hand retains flat finger geometry, whereas the No-Skill hand introduces distal finger droop that narrows the capture basin.

\begin{figure}[b]
  \centering
  \includegraphics[width=\columnwidth]{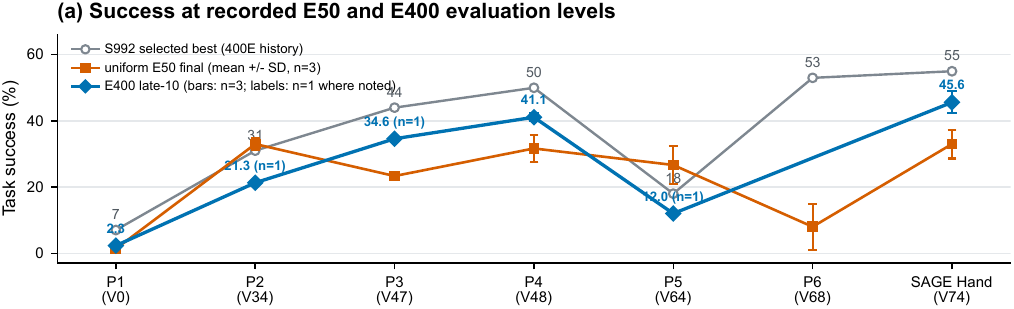}\\[2pt]
  \includegraphics[width=\columnwidth]{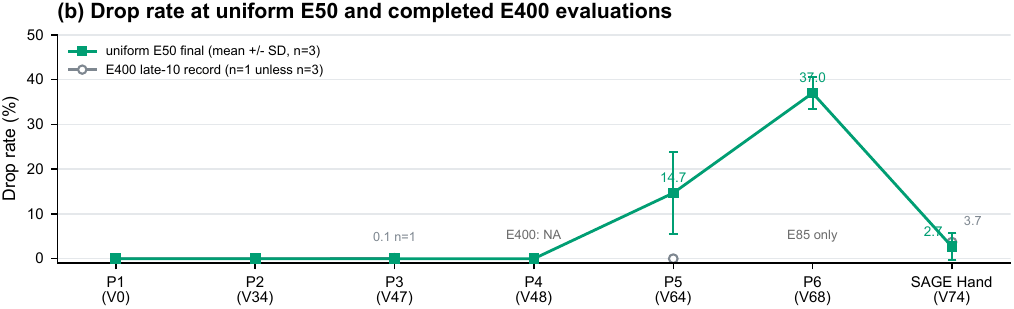}
 \caption{\textbf{Design history of SAGE Hand.}
\textbf{(a)}~Gray circles show historical best success from seed-992 400-epoch runs; orange squares show uniform 50-epoch final success (mean$\pm$s.d., $n=3$); blue diamonds show E400 late-10 success, the primary late-training metric. Bars denote between-seed s.d.; labels mark $n=1$ where applicable.
\textbf{(b)}~Green squares show 50-epoch drop rate and gray circles show available E400 late-10 drop records. The abscissa follows design-version order.}
  \label{fig:history}
\end{figure}

\subsection{Morphology Evolution under the Interactive Loop}
\noindent Under the common asymmetric-hexagon protocol, late-training success rises from $2.3\pm0.3$\% at Prototype~P1 to $41.1\pm1.2$\% at Prototype~P4 (V48) and $45.6\pm3.3$\% at SAGE Hand (mean$\pm$between-seed SD, Fig.~\ref{fig:history}). The Prototype~P1$\rightarrow$P4 increase coincides with reset correction and replacement of the P1 position-actuator adapter by a trainable action interface; the P4$\rightarrow$SAGE Hand gain follows from the integrated palm, compact roots, tip, and thumb revisions.

\begin{table*}[t!]
\captionsetup{font=small}
\caption{Task success, mean$\pm$between-seed SD over three seeds.}
\label{tab:results}
\centering
\footnotesize
\begin{tabular}{lcccc}
\toprule
Object & SAGE Hand no tactile & SAGE Hand T4 tactile & LEAP hand & Shadow hand \\
\midrule
Asymmetric hexagonal prism & $45.6\pm3.3$\% (3) & $45.4\pm2.2$\% (3) & $37.3\pm2.5$\% (3) & $15.8\pm1.3$\% (3) \\
Colored cube & $42.6\pm4.0$\% (3) & $43.4\pm0.5$\% (3) & $10.5\pm1.6$\% (3) & $0.0$\% (1) \\
Bicolor cylinder & $70.8\pm5.6$\% (3) & $80.1\pm5.9$\% (3) & $51.6\pm1.1$\% (3) & $18.9\pm4.8$\% (3) \\
\bottomrule
\end{tabular}
\end{table*}

The uniform 50-epoch screens reached 34\%, 23\%, 31\%, 9\%, and 13\% at Prototype~P2, Prototype~P3, Prototype~P4, V67, and Prototype~P5 (seed 992, 100 evaluation episodes at epoch 50). Their best observed checkpoints were 42\% at epoch 25, then 32\%, 31\%, 20\%, and 17\% at epochs 45, 50, 45, and 45. The version index alone does not predict final capability; the frozen three-seed endpoint is reported below.

\begin{figure*}[!t]
  \centering
  \includegraphics[width=0.86\textwidth]{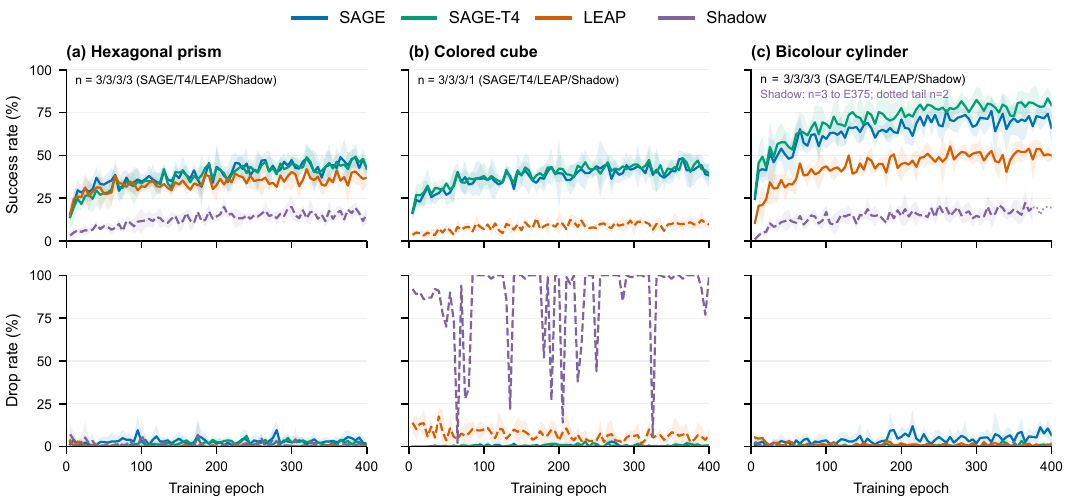}
  \caption{\textbf{Training-time success (top) and drop (bottom) rates} for SAGE Hand, SAGE Hand-T4, LEAP, and Shadow on (a) the hexagonal prism, (b) the colored cube, and (c) the bicolor cylinder, under the same nominal $400$-epoch budget and one frozen learning protocol. Lines are seed means, bands $\pm1$~s.d.; seed counts appear in each panel ($n=1$ for Shadow on the cube; $n=3\to2$ after epoch $375$ for Shadow on the cylinder). Each point reports a teacher-free evaluation of the checkpoint saved at that epoch. The primary summary statistic is the mean over the final ten available evaluation points.}
  \label{fig:training}
\end{figure*}

\subsection{Three-Object Performance}
\noindent Table~\ref{tab:results} gives the completed cells and Fig.~\ref{fig:training} shows the corresponding training curves. On the asymmetric hexagon SAGE Hand reaches $45.6$\% task success against $37.3$\% for the LEAP hand, a gap of $8.3$ percentage points (95\% CI $[1.5,15.1]$, $p=0.03$). SAGE Hand-T4 reaches $45.4$\% over three seeds, indistinguishable from SAGE Hand without tactile input ($-0.2$ points, CI $[-6.9,6.5]$). 

\begin{figure*}[t!]
  \centering
  \includegraphics[width=0.85\textwidth]{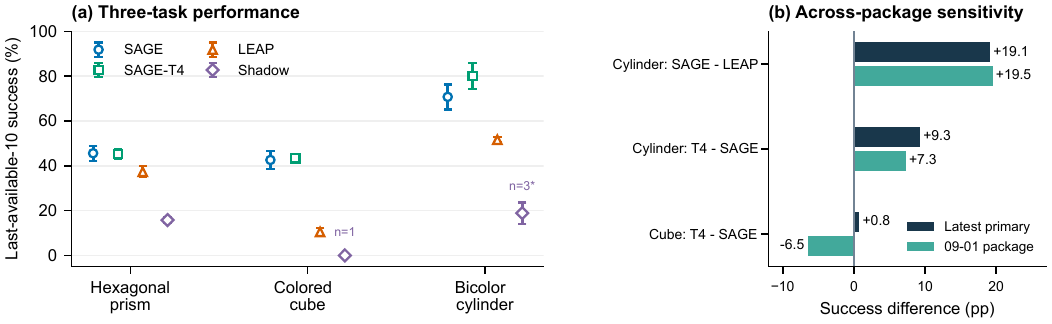}
  \caption{\textbf{Summary of the frozen-protocol evaluation.} (a)~Last-available-10 success for four hands on three objects; markers are seed means, bars $\pm1$~s.d., $n=3$ except where annotated ($n=1$ for Shadow on the cube, $n=3^{*}$ for Shadow on the cylinder). (b)~The same pairwise differences were recomputed on two evaluation packages. Both cylinder comparisons hold their sign across packages; the cube comparison reverses ($+0.8$ vs $-6.5$~pp) and is left undetermined.}
  \label{fig:results}
\end{figure*}

On the cylinder, SAGE Hand reaches $70.8\pm5.6$\% against $51.6\pm1.1$\% for LEAP in the primary evaluation package, a $19.1$-point advantage (95\% CI $[5.9,32.3]$, $p=0.024$). The 09-01 package shows a similar gap of $19.5$ points ($69.9$\% vs.\ $50.4$\%). Thus, the sign and approximate magnitude of the SAGE--LEAP cylinder advantage persist across changes in replay capacity and hardware (Fig.~\ref{fig:results}).

\subsection{Object-Dependent Tactile Effect}
\noindent Tactile observations increase cylinder success from $69.9$\% to $77.2$\% in 09-01 and from $70.8$\% to $80.1$\% in the primary evaluation package, gains of $+7.3$ and $+9.3$ points; both confidence intervals include zero. More consistently, cylinder drop rate decreases from $2.87$\% to $1.17$\% and from $7.23$\% to $1.00$\%, respectively. Thus, dense temporal tactile input most clearly improves retention. The cube tactile effect changes sign between primary package ($+0.8$ points) and 09-01 package($-6.5$ points), with both confidence intervals including zero; cube-rotation success is therefore unresolved.

\section{Discussion}

\noindent \textbf{RQ1}: the gates reject plausible but physically inadmissible candidates, and the fixed-budget audit places 203 of 209 physics-trial failures at the signed-motion gate, characterizing the boundary of full autonomy. The controlled iteration study confirms that the gates also function as a regression guard: the Skill path preserves capture at 164/165 while the path without the Skill degrades to 14/165. \textbf{RQ2}: The matched SAGE--LEAP comparison isolates task specialization on matched action spaces on the asymmetric hexagon ($+8.3$ points, interval clear of zero) and on the bicolor cylinder in the primary evaluation package ($+19.1$ points, interval clear of zero); the 09-01 package reproduces the cylinder advantage in sign and approximate magnitude. \textbf{RQ3}: The cylinder advantage retains its sign across both packages. Temporal tactile input consistently improves cylinder retention.

The design trajectory, the SAGE Hand--LEAP comparison, and the T4 ablation operate at different levels of evidence. The trajectory is a supervised design path: each local decision traces to a physical blocker, but the endpoint reflects all 74 revisions jointly. The SAGE Hand--LEAP comparison compares task specialization on matched action spaces; the T4 ablation isolates sensing on identical collision geometry.

Of 165 deterministic closure controls, 14 produced functional, nonpenetrating captures ($8.5$\%). The minimum signed distance was $0.102$~mm for SAGE Hand and $1.908$~mm for LEAP. Across $3{,}000$ random actions, the strict guard intervened 516 times for SAGE Hand and 22 times for LEAP, with no numerical failures. Together, the smaller clearance and higher intervention count are consistent with SAGE Hand operating closer to the contact-validity boundary.

\textit{Future work.} Physical deployment is the natural next step: the current MuJoCo model omits manufacturing tolerance, backlash, cable compliance, actuator thermal limits, and noise, so sim-to-real transfer will test whether the reported advantage holds on hardware. Direction-resolved evaluation ($+Z$ vs $-Z$) would further clarify if the gain is bidirectionally balanced in the sign the thumb geometry favors.

\section{Conclusion}
\noindent We presented \skillname{}, a contact-grounded Skill for parameterized dexterous-hand design. It turns a functional brief into executable robot descriptions, rejects invalid candidates at physical gates ordered by cost, and uses a fixed RL protocol to evaluate what survives. That loop produced SAGE Hand, a four-finger 16-DoF hand that reaches $45.6$\% task success on an asymmetric hexagonal prism against $37.3$\% for LEAP under matched conditions, and 70.8\% against 51.6\% on a bicolor cylinder, with tactile input more than halving the drop rate. The Skill path preserves capture at 164/165 trials while the path without the Skill degrades to 14/165, confirming that the gates act as a regression guard. Each accepted candidate ships with its models, gate records, checkpoints, and provenance.

\bibliographystyle{IEEEtran}
\bibliography{references}
\end{document}